\documentclass[preprint,12pt]{elsarticle}

\usepackage{amssymb}
\usepackage{amsmath}

\usepackage{booktabs}
\usepackage{multirow}
\usepackage{multicol}
\usepackage{acronym}
\usepackage{todonotes}
\usepackage{adjustbox}
\usepackage{arydshln}
\usepackage{hyperref}

\acrodef{AMNLT}{Aligned Music Notation and Lyrics Transcription}
\acrodef{OMR}{Optical Music Recognition}
\acrodef{OCR}{Optical Character Recognition}
\acrodef{HTR}{Handwritten Text Recognition}
\acrodef{CRNN}{Convolutional Recurrent Neural Netwwork}
\acrodef{CNN}{Convolutional Neural Network}
\acrodef{RNN}{Recurrent Neural Network}
\acrodef{CTC}{Connectionist Temporal Classification}
\acrodef{BLSTM}{Bidirectional Long-Short Term Memory}
\acrodef{FCN}{Fully Convolutional Network}
\acrodef{DSC}{Deep Separable Convolutional}
\acrodef{CNNT2D}{Convolutional Neural Network with Transformer 2D}
\acrodef{SMT}{Sheet Music Transformer}
\acrodef{MEI}{Music Encoding Initiative}
\acrodef{DDMAL}{Distributed Digital Music Archives \& Libraries Lab}
\acrodef{SMT}{Sheet Music Transformer}

\newcommand{\GregoSynth}{\textsc{GregoSynth}}
\newcommand{\Solesmes}{\textsc{Solesmes}}
\newcommand{\Einsiedeln}{\textsc{Einsiedeln}}
\newcommand{\Salzinnes}{\textsc{Salzinnes}}
\newcommand{\gabc}{\textsc{gabc}}           
\newcommand{\sgabc}{\textsc{s-gabc}}        
\newcommand{\pgabc}{\textsc{pseudo gabc}}   

\newcommand{\CER}{\textsc{CER}}     
\newcommand{\MER}{\textsc{MER}}     
\newcommand{\SylER}{\textsc{SylER}} 
\newcommand{\AMLER}{\textsc{AMLER}} 
\newcommand{\bWER}{\textsc{bWER}}   
\newcommand{\AlER}{\textsc{AlER}}   

\makeatletter
\def\adl@drawiv#1#2#3{%
        \hskip.5\tabcolsep
        \xleaders#3{#2.5\@tempdimb #1{1}#2.5\@tempdimb}%
                #2\z@ plus1fil minus1fil\relax
        \hskip.5\tabcolsep}
\newcommand{\cdashlinelr}[1]{%
  \noalign{\vskip\aboverulesep
           \global\let\@dashdrawstore\adl@draw
           \global\let\adl@draw\adl@drawiv}
  \cdashline{#1}
  \noalign{\global\let\adl@draw\@dashdrawstore
           \vskip\belowrulesep}}
\makeatother

\graphicspath{{./img/}}

\journal{Pattern Recognition}

\begin{document}

\begin{frontmatter}



\title{Aligned Music Notation and Lyrics Transcription}


\author[1]{Eliseo Fuentes-Martínez\corref{cor1}}
\cortext[cor1]{Corresponding Author}
\ead{eliseo.fuentes@ua.es}
\author[1]{Antonio Ríos-Vila} 
\ead{arios@dlsi.ua.es}
\author[1]{Juan C. Martinez-Sevilla} 
\ead{jcmartinez.sevilla@ua.es}
\author[1,2]{David Rizo} 
\ead{drizo@dlsi.ua.es}
\author[1]{Jorge Calvo-Zaragoza} 
\ead{jcalvo@dlsi.ua.es}

\affiliation[1]{organization={Pattern Recognition and Artificial Intelligence Group, University of Alicante},
            city={Alicante},
            country={Spain}}
            
\affiliation[2]{organization={Instituto Superior de Enseñanzas Artísticas de la Comunidad Valenciana},
            city={Alicante},
            country={Spain}}

\begin{abstract}
The digitization of vocal music scores presents unique challenges that go beyond traditional Optical Music Recognition (OMR) and Optical Character Recognition (OCR), as it necessitates preserving the critical alignment between music notation and lyrics. This alignment is essential for proper interpretation and processing in practical applications. This paper introduces and formalizes, for the first time, the Aligned Music Notation and Lyrics Transcription (AMNLT) challenge, which addresses the complete transcription of vocal scores by jointly considering music symbols, lyrics, and their synchronization. We analyze different approaches to address this challenge, ranging from traditional divide-and-conquer methods that handle music and lyrics separately, to novel end-to-end solutions including direct transcription, unfolding mechanisms, and language modeling. To evaluate these methods, we introduce four datasets of Gregorian chants, comprising both real and synthetic sources, along with custom metrics specifically designed to assess both transcription and alignment accuracy. Our experimental results demonstrate that end-to-end approaches generally outperform heuristic methods in the alignment challenge, with language models showing particular promise in scenarios where sufficient training data is available. This work establishes the first comprehensive framework for AMNLT, providing both theoretical foundations and practical solutions for preserving and digitizing vocal music heritage.
\end{abstract}




\end{frontmatter}




\section{Introduction}
The main obstacle to carrying out digital musicology tasks on a large scale is the transcription of written musical sources into a format that can be further processed by a computer \cite{Me2016}. This transcription process is costly when done manually, as the complexity of music notation requires the use of specialized and hard-to-manage music score editors, along with expert supervision. The challenge becomes even more discouraging for historical music notation systems, for which suitable tools might not exist. Consequently, automatic transcription systems for music documents are invaluable~\cite{Alfaro_Contreras:ISMIR:2021}.

\ac{OMR} is a field of computer science dedicated to reading music notation from document images \cite{CalvoZaragoza:ACMCS:2020}. It has been an active research area for decades \cite{CalvoZaragoza2023}. Typically, the output of an OMR system is a structured digital format, such as MusicXML or MEI, which encodes the musical content for further processing.

\begin{figure}[h]
    \centering
    \includegraphics[width=0.8\linewidth]{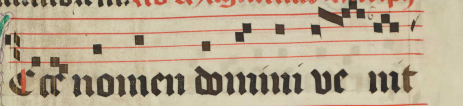}
    \caption{Vocal music excerpt. It is important to mention how, in this type of music, there is no one-to-one relationship between musical notes and lyrics; it is rather a many-to-many relationship, thus making the alignment so crucial to know how these scores should be interpreted.}
    \label{fig:excerpt}
\end{figure}

Traditionally, OMR systems focused on the detection and recognition of music symbols using heuristic image processing techniques \cite{rebelo2012optical}. However, deep learning brought about a paradigm shift \cite{tuggener2024real,mayer2024practical,yesilkanat2024full}, opening new possibilities to advance the field that were once considered infeasible. One such task is the automatic transcription of vocal music documents. Vocal music refers to compositions where the singing part is central to the piece, whether accompanied by instruments or not. Thus, an OMR system for this type of music must handle not only the transcription of the music notation but also the lyrics that indicate the words to be sung. Both modalities represent complementary aspects of the same musical work: the text specifies ``what'' to sing, while the music notation specifies ``how'' to sing it (see Fig.~\ref{fig:excerpt}).

While \ac{OMR} algorithms can independently recognize the music notation \cite{Baro:PRL:2019} and text recognition algorithms can handle the lyrics \cite{li2025htr}, such approaches fail to address the alignment between notes and lyrics---a critical requirement for meaningful musicological outcomes. This challenge is particularly interesting from a scientific perspective, as there is no existing deep learning framework for it.

\begin{figure}[h]
    \centering
    \includegraphics[width=0.8\linewidth]{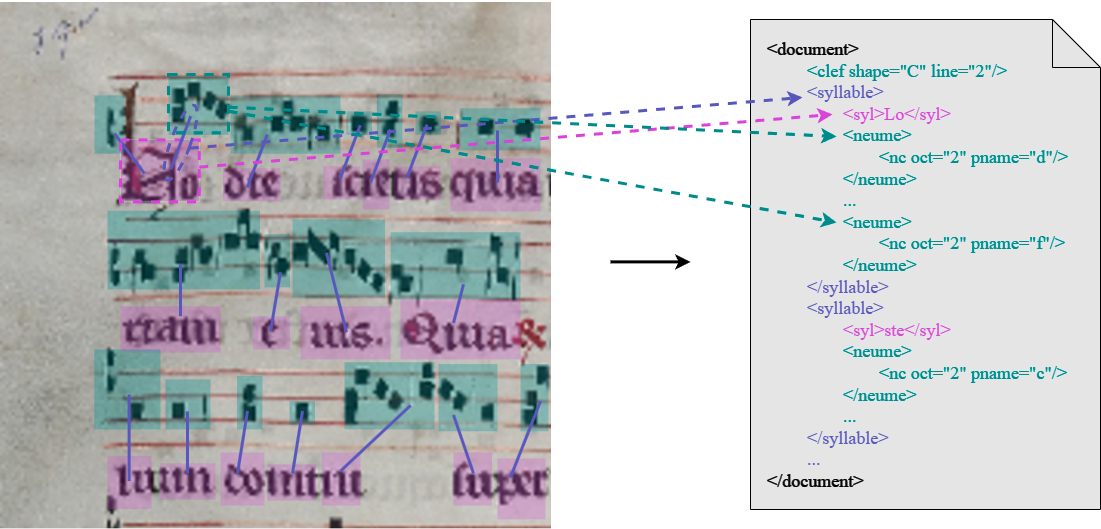}
    \caption{Graphical representation of the AMNLT framework, depicting its three key components: music notation (turquoise), lyrics (magenta), and their alignment (purple).}
    \label{fig:amnlt_scheme}
\end{figure}

This paper is the first in the field of \ac{OMR} to comprehensively address the transcription of vocal scores. We achieve this by formally defining the \ac{AMNLT} challenge, which emphasizes not only the transcription of music notation and lyrics but also the critical alignment task (see Fig.~\ref{fig:amnlt_scheme}). We also analyze existing divide-and-conquer approaches and propose how end-to-end solutions can be adapted to include alignment information for music scores. Additionally, this paper introduces a set of metrics to assess the performance, evaluating both transcription and alignment accuracy. All these aspects are evaluated through experimentation on four AMNLT scenarios, comprising three real datasets and one hybrid dataset. The results demonstrate that (i) divide-and-conquer methods, while precise in transcription, fail to provide complete results for AMNLT, and (ii) the end-to-end approaches outperform traditional methods in both transcription quality and alignment precision, establishing a new baseline for full AMNLT.

The remainder of the paper is structured as follows: Section \ref{sec:relworks} reviews how the OMR literature addresses the transcription of music and lyrics, highlighting overlooked aspects. Section \ref{sec:amnlt} provides a formal definition of the AMNLT challenge. Section \ref{sec:amnlt_approaches} discusses the adaptations required for state-of-the-art transcription pipelines to address AMNLT and introduces two holistic methods to address this challenge. In Section \ref{sec:case_study}, we describe the case study and present the four datasets used in our experiments, while Section \ref{sec:metrics} explains the metrics developed to evaluate the performance of the models. Section \ref{sec:amnlt_approaches_implementation} details the implementation of the proposed approaches. The results are discussed in Section \ref{sec:results}, and the paper concludes in Section \ref{sec:conclusions}.

\section{Background}
\label{sec:relworks}
\ac{OMR} methods typically begin with layout analysis, where the document is divided into distinct regions of interest such as staves, lyrics, titles, and other elements. This step is essential for isolating the structural components of a music score. Current layout analysis methods for music scores are capable of robust and accurate region extraction, even under challenging conditions \cite{waloschek2019identification,Castellanos:ESWA:2022}.

Once the regions are extracted, modern systems predominantly adopt end-to-end methods to retrieve the content within each region in a single step \cite{CastellanosCI20}. This formulation offers significant advantages over traditional symbol-based pipelines by learning contextual relationships directly from data. Typically, these methods are based on Convolutional Recurrent Neural Networks (CRNN) combined with \ac{CTC} loss \cite{Calvo-Zaragoza:PRL:2019,Alfaro-Contreras22,mayer2024practical}, although some authors have recently incorporated the use of Transformers \cite{smt_toto}. This end-to-end approach has achieved notable success across various types of musical sources, including both printed and handwritten documents, and can be considered the state of the art for \ac{OMR}.

For tasks involving vocal music, the transcription challenge extends beyond recognizing music notation to include lyrics transcription, which is traditionally handled independently \cite{Burgoyne:ISMIR:2009}. While it is possible to reliably extract music and lyrics information separately, this approach fails to address the critical alignment between the two modalities. As mentioned above, this alignment is essential for interpreting vocal music, as it synchronizes the ``what'' (lyrics) with the ``how'' (music notation). This concept of alignment has been widely studied in other domains, such as the automatic transcription of music from audio recordings \cite{Gupta:ICASSP:2020, Stoller:ICASSP:2019, Sharma:ICASSP:2019}, yet it remains largely unexplored in the context of document image analysis for vocal music.

The literature reveals only a few attempts to address the interplay between music and lyrics. Villarreal et al. \cite{Villarreal:DOCENG:2023, Villarreal:ICDAR:2024} introduced approaches that acknowledge this interaction, leveraging it to improve the independent performance of OMR and OCR methods. However, their work continues the paradigm of treating the two tasks separately, without addressing alignment as a core challenge. Martinez-Sevilla et al. \cite{MartinezSevilla:ICDAR:2023} represent the first attempt to directly address the aligned transcription of vocal music. Their approach demonstrates the feasibility of producing aligned outputs but relies solely on synthetic data, limiting its applicability to real-world scores. Moreover, their work does not propose a general formulation of the problem or define metrics for evaluating transcription and alignment quality comprehensively.

In addition to academic research, there are also practical tools such as OMMR4ALL \cite{OMMR4ALL} and the Cantus Analysis Tool \cite{Cantus}, which attempt to align music and lyrics using heuristic-based systems. These tools primarily rely on object detection and handcrafted rules, offering solutions tailored to specific datasets or use cases rather than generalizable methods. As such, they do not provide a robust framework or benchmarks for alignment.

Consequently, existing literature falls short of providing a unified and comprehensive approach to fully transcribing vocal music scores. This paper addresses this gap by introducing the \ac{AMNLT} framework, which integrates alignment into the transcription process. Unlike prior studies, our work formalizes the problem, introduces suitable evaluation metrics, and benchmarks the methods using diverse datasets, including real sources.

\section{Aligned Music Notation and Lyrics Transcription (AMNLT)}
\label{sec:amnlt}
The \ac{AMNLT} challenge focuses specifically on vocal music scores. Let us denote $\mathcal{X}$ as the space of (vocal) music score images and $\mathcal{Y}$ as the corresponding content transcription space. Each $y \in \mathcal{Y}$ comprises two underlying languages: music notation and lyrics.

Lyrics are a unique component of the music score, as they represent text with a direct musical function. As such, lyrics can be understood as a distinct voice within the music score. While lyrics are encoded separately from the rest of the musical symbols, they share the same overarching musical meaning. This distinction means that the two sources---music and lyrics---originate from different domains: music encoding ($\Sigma_m$) and natural language ($\Sigma_l$)\footnote{Note that this formulation is not tied to any specific language or music encoding.}. Despite their distinct domains, lyrics are inherently dependent on music notation, as their interpretation is intrinsically linked to the notes. Specifically, the notes determine the pitch at which each syllable is performed. This dependency is essential; without it, the interpretation of vocal music would be meaningless. This critical relationship is formalized as the \textit{alignment} between music and lyrics (Fig.~\ref{fig:amnlt_scheme}).

Given a sequence of music notation elements $M = (m_1, m_2, ..., m_n)$ and a sequence of syllables $L = (l_1, l_2, ..., l_n)$, we define the alignment as a partitioning of the music elements into disjoint groups $A_1, A_2, ..., A_n$, where each group $A_j$ corresponds to a specific syllable $l_j$. Each $A_j \subseteq M$ satisfies $A_i \cap A_j = \emptyset$ for all $i \neq j$ and $M = A_1 \cup A_2 \cup ... \cup A_n$. Furthermore, every syllable $l_j$ must have at least one associated music group, so $A_j \neq \emptyset ~\forall j$. This relationship is formalized by an underlying alignment function $a: L \rightarrow \mathcal{P}(M)$, such that $a(l_j) = A_j$.

Given $x \in \mathcal{X}$, the transcription challenge can be first approximated by seeking $\hat{y} \in \mathcal{Y}$ such that:

\begin{equation}
\label{eq:seqret}
\hat{\mathbf{y}} = \arg\max_{\mathbf{y} \in \Sigma} P(\mathbf{y} \mid x)
\end{equation}

\noindent where $\Sigma$ denotes the vocabulary for vocal music transcription. Since the problem involves multimodal outputs, Eq.~\ref{eq:seqret} can be further decomposed into two objectives: one for music (Eq.~\ref{eq:OMR}) and another for lyrics (Eq.~\ref{eq:OCR}).

\begin{equation}
\label{eq:OMR}
\mathbf{\hat{y}_m} = \arg\max_{\mathbf{y_m} \in \Sigma_{m}} P(\mathbf{y_m} \mid x)
\end{equation}

\begin{equation}
\label{eq:OCR}
\mathbf{\hat{y}_l} = \arg\max_{\mathbf{y_l} \in \Sigma_{l}} P(\mathbf{y_l} \mid x)
\end{equation}

These formulae represent the independent challenges of \ac{OMR} and \ac{OCR}. However, in the context of vocal music, these outputs must eventually be aligned. To achieve this, we can estimate the most probable alignment between the music notation sequence and the lyrics sequence:

\begin{equation}
\label{eq:alignment_approximation}
    \mathcal{\hat{A}}(\mathbf{\hat{y}_l}, \mathbf{\hat{y}_m}) \underset{\hat{y}_l \in \Sigma_{l},  \hat{y}_m \in \Sigma_{m}}{=} \max P(A_{\hat{y}_m} \mid \hat{y}_l)
\end{equation}

Thus, the \ac{AMNLT} task can be defined as estimating the most probable aligned sequence of music notation and lyrics from the given input image:

\begin{equation}
\label{eq:amnlt_5}
    \mathbf{\hat{y}} = \arg\max P(\mathcal{\hat{A}}(\mathbf{\hat{y}_l}, \mathbf{\hat{y}_m}) \mid x)
\end{equation}

Note that it is the alignment process that gives the problem its full meaning, as the task involves not only transcribing the image content but also determining the relationships between the musical and textual elements of the scores---either by following predefined rules or allowing an end-to-end model to infer them.

\section{Approaches for AMNLT}
\label{sec:amnlt_approaches}
In this section, we analyze how state-of-the-art OMR and OCR transcription methods can be combined to address \ac{AMNLT}. We also propose several end-to-end approaches as alternatives, which directly provide an aligned output. 

For the purposes of this paper, we will assume that a prior layout analysis step has been performed to extract isolated regions containing a single system (a group of one staff and its corresponding lyrics line). This step can be effectively accomplished using existing methods, as discussed in Section~\ref{sec:relworks}.

The approaches discussed here are broadly organized into two categories: post-alignment methods, here referred to as \emph{divide \& conquer}, and holistic methods.

\subsection{Divide \& Conquer}
\label{sec:d&c}
\begin{figure}[h]
    \centering
    \includegraphics[width=\linewidth]{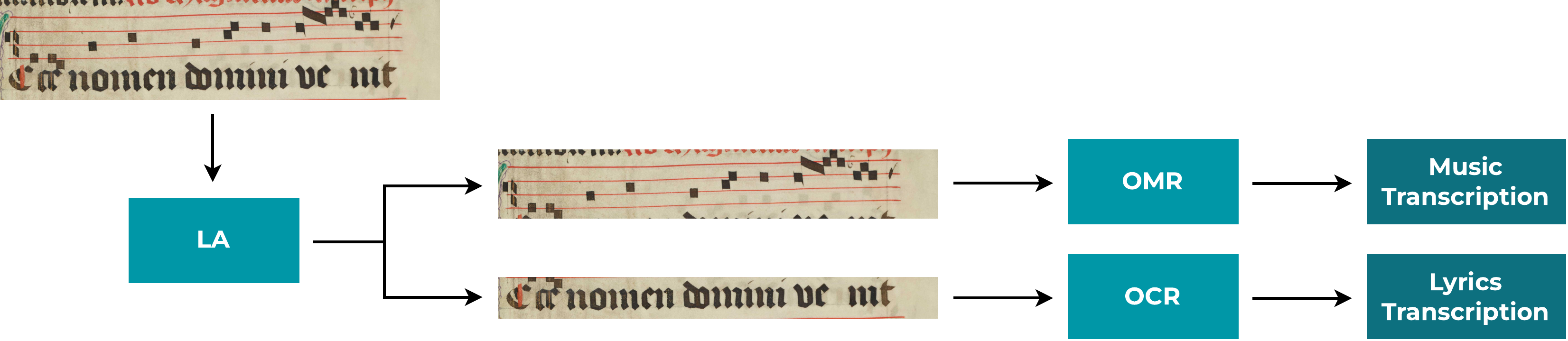}
    \caption{Graphical example of the \emph{divide \& conquer} approach. The system is split into two parts: on one hand, the music is transcribed by OMR methods, and on the other, the lyrics are transcribed by OCR methods. Then, an ad-hoc post-process must be considered to align both output modalities.}
    \label{fig:d&c}
\end{figure}

The first approach to analyze is the \emph{divide \& conquer} strategy. In this approach, two independent \ac{OMR} and \ac{OCR} models are used to transcribe their respective content from the input image, addressing Eq.~\ref{eq:OMR} and Eq.~\ref{eq:OCR} individually. This process is illustrated in Fig.~\ref{fig:d&c}. Both \ac{OMR} and \ac{OCR} methods are assumed to be trained using the framework provided by \ac{CTC}, consistent with state-of-the-art approaches~\cite{CalvoZaragoza2018,Calvo-Zaragoza:PRL:2019,Baro:PRL:2019,Puigcerver:ICDAR:2017,Coquenet:ICFHR:2020}. 

To approximate the alignment function that relates the outputs of both networks, as defined in formula~\ref{eq:alignment_approximation}, a post-processing step is required. In this work, we consider two post-processing methods: (i) syllable-level post-alignment and (ii) frame-level post-alignment.

\subsubsection{Syllable-level post-alignment}
The first post-alignment strategy follows a greedy approach, where the generated syllables are paired with the obtained musical groups one by one: each lyrics syllable from the lyrics transcription is coupled with the musical group in the corresponding (ordinal) position in the music transcription. This method leverages the fact that, when dividing the transcriptions into musical and textual parts, each fragment is naturally separated into character groups (syllables for text and musical groups, associated with each syllable, for music). Note that, to enable this method, it is necessary that the models are trained, using a properly annotated ground-truth, to produce sequences with separations between the groups of each modality. As a result, syllable-based alignment becomes a natural step.

The alignment is carried out as follows: (i) predictions from each model are stored, ensuring they are saved as aligned pairs, and (ii) both files are processed simultaneously, pairing each musical group with the corresponding syllable. Unmatched groups are concatenated in the output without pairing. For instance, given a lyrics transcription $\mathbf{\hat{y}_l} = (s_1, s_2, s_3, ..., s_n)$ and a music transcription $\mathbf{\hat{y}_m} = (m_1, m_2, m_3, ..., m_n)$---where $s$ refers to a lyrics syllable and $m$ refers to a music group---the resulting aligned sequence would be $\mathcal{A}(\mathbf{\hat{y}_l}, \mathbf{\hat{y}_m}) = ((s_1, m_1), (s_2, m_2), (s_3, m_3), ..., (s_n, m_n))$. 

Note that this approach is potentially suboptimal, as errors in musical grouping by the \ac{OMR} method can easily propagate and lead to misalignment issues.

\begin{figure}[h]
    \centering
    \includegraphics[width=\linewidth]{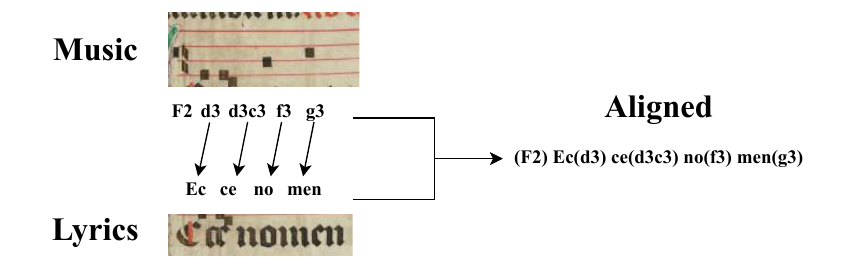}
    \caption{Example of the \emph{syllable-level post-alignment}, where each music group from the music transcription is paired with the lyrics syllable in the corresponding position in the lyrics transcription.}
    \label{fig:syl-lvl_post-alignment}
\end{figure}

\subsubsection{Frame-level post-alignment}
\label{para:frame}
This method leverages the \ac{CTC} training strategy output. In this approach, music and lyrics transcriptions are aligned using the Viterbi alignment algorithm~\cite{Villarreal:DOCENG:2023}. For each frame containing a non-empty character, the algorithm identifies the nearest non-empty character frame from the complementary \ac{CTC} sequence, either to the left or right. By aligning frames in this manner, a complete \ac{AMNLT} transcription is produced. 

However, this method requires that both posteriorgrams have the same number of frames. Consequently, the input images for music notation and lyrics must have identical widths for this approach to function correctly.

\begin{figure}[h]
    \centering
    \includegraphics[width=\linewidth]{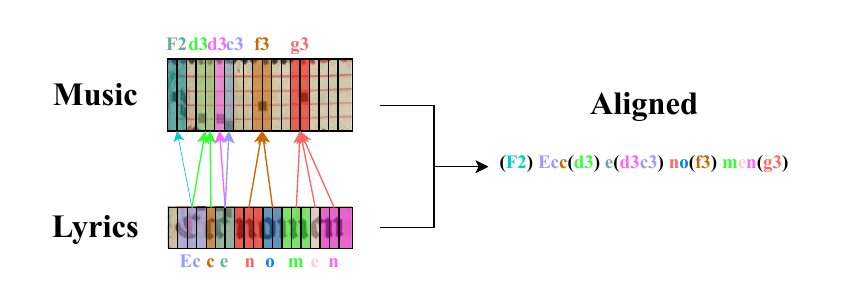}
    \caption{Example of the \emph{frame-level post-alignment}, where each lyrics frame is coupled with its nearest musical frame. Due to this frame-wise post-alignment and the inherent graphical misalignment between music and lyrics for space reasons, some errors may occur. For instance, as shown, the brown \emph{c} is incorrectly aligned with the first syllable instead of the second, where it should belong.}
    \label{fig:frame-lvl_post-alignment}
\end{figure}

\subsection{Holistic methods}
\label{sec:holistic_methods}
In this work, we propose holistic methods as an alternative to post-alignment strategies. End-to-end approaches must address \ac{AMNLT} by directly transcribing and aligning the content of the input score through a single model. To achieve this, we integrate all tasks into the output vocabulary of the model. Specifically, we combine three different sets: the music notation vocabulary ($\Sigma_m$), the lyrics character set ($\Sigma_l$), and an alignment vocabulary ($\mathcal{\hat{A}}$) that relates $\Sigma_m$ and $\Sigma_l$. This integration enables a model to approximate both the transcription and alignment functions in a single step, directly solving Eq.~\ref{eq:amnlt_5}. 

\subsubsection{Base holistic approach}
\label{sec:base_holistic}
The first approach involves the direct application of a vocabulary-based strategy to the end-to-end methods described in Section~\ref{sec:d&c}. Instead of creating separate models for each task and combining their outputs through post-processing, this approach produces a complete \ac{AMNLT} output using a single end-to-end model. In other words, a single model is considered for which the output vocabulary naturally integrates music notation, lyrics, and their alignment.

By leveraging the same architectures and frameworks as the state of the art, this method provides a baseline for evaluating holistic approaches.

\subsubsection{Unfolding}
\label{sec:unfolding}
The \emph{unfolding} approach builds on recent advances in the \ac{HTR} and \ac{OMR} fields. This method is devised to achieve a better implicit alignment between the source image and its transcription~\cite{Yousef:CVPR:2020,Coquenet2021,RiosVila:IJDAR:2023}. Instead of processing input features as a sequence from left to right,\footnote{This strategy typically involves a vertical collapse of the feature maps to process them as a sequence.} these approaches learn to sequentially read the unfolded feature map derived from the input image. This enables the model to process the document content in the same reading order as the ground truth annotation.

\begin{figure}[h]
    \centering
    \includegraphics[width=.5\linewidth]{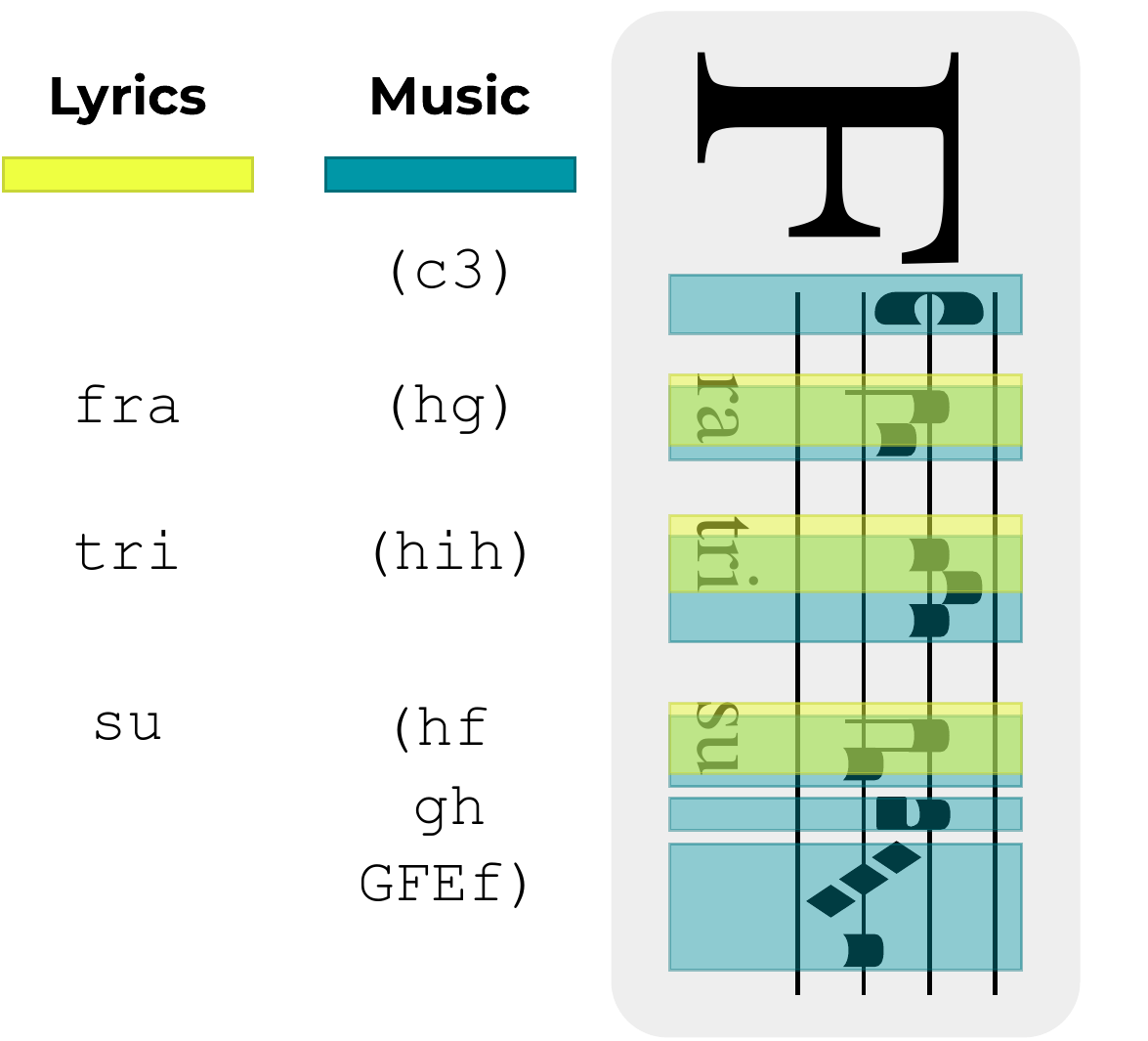}
    \caption{Fragment of a Gregorian chant with alignment information in \gabc{} format. Colored boxes indicate pairs of lyrics and music. Although the image shows a specific encoding of the score, this structure can also be found in other standard music encoding formats.}
    \label{fig:unfolding_rotation}
\end{figure}

\begin{figure}[h]
    \centering
    \includegraphics[width=\linewidth]{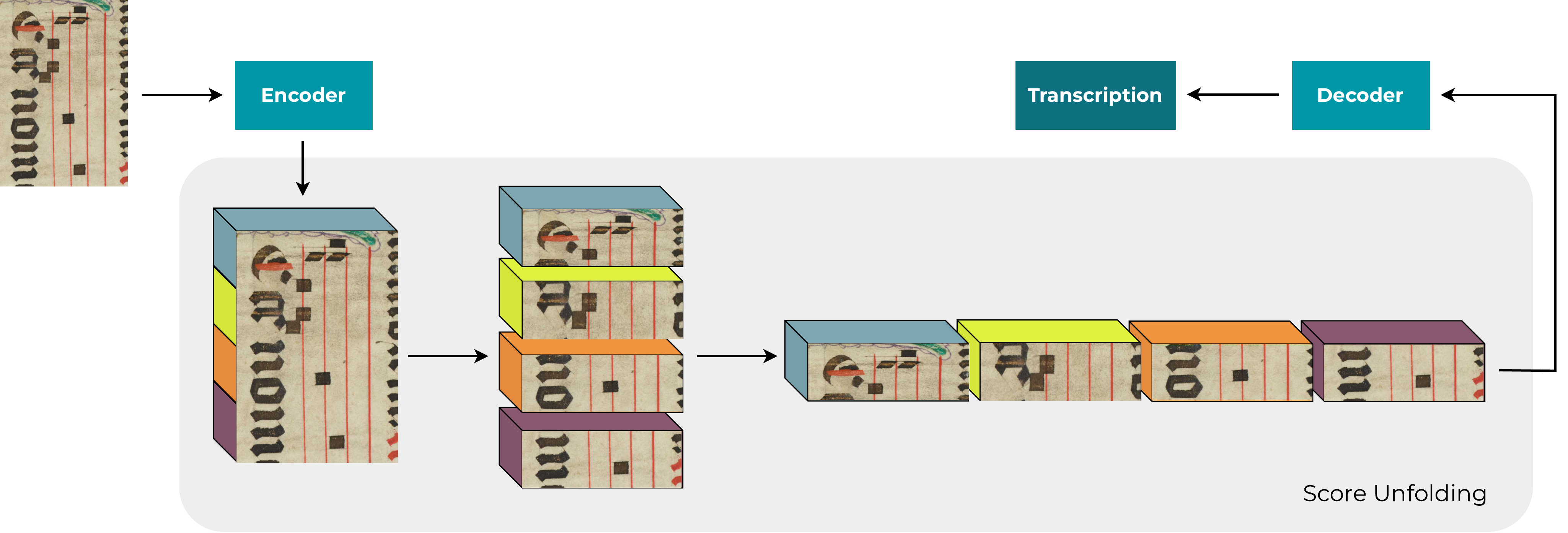}
    \caption{Visualization of the unfolding mechanism for transcribing \ac{AMNLT} scores.}
    \label{fig:unfolding_stave}
\end{figure}

For vocal music scores, the lyrics are graphically closer to their corresponding music notes. This idea has been preliminary studied in the work of Martinez-Sevilla et al.~\cite{MartinezSevilla:ICDAR:2023}, where authors propose rotating the music system clockwise to achieve a graphical alignment between music notes and lyrics that matches the ground truth. Figures~\ref{fig:unfolding_rotation} and \ref{fig:unfolding_stave} provide visual examples of this approach. 

In this work, we also consider the unfolding method for addressing \ac{AMNLT} in an end-to-end fashion.

\subsubsection{Language modeling}
\label{sec:method_language_modeling}
The holistic methods described so far are assumed to be trained using \ac{CTC}. However, \ac{CTC}-based methods are known to be limited by the sequential nature of image features. For instance, unfolding mechanisms require additional processing steps to \textit{align} image features with the ground truth structure.

\begin{figure}[h]
    \centering
    \includegraphics[width=\linewidth]{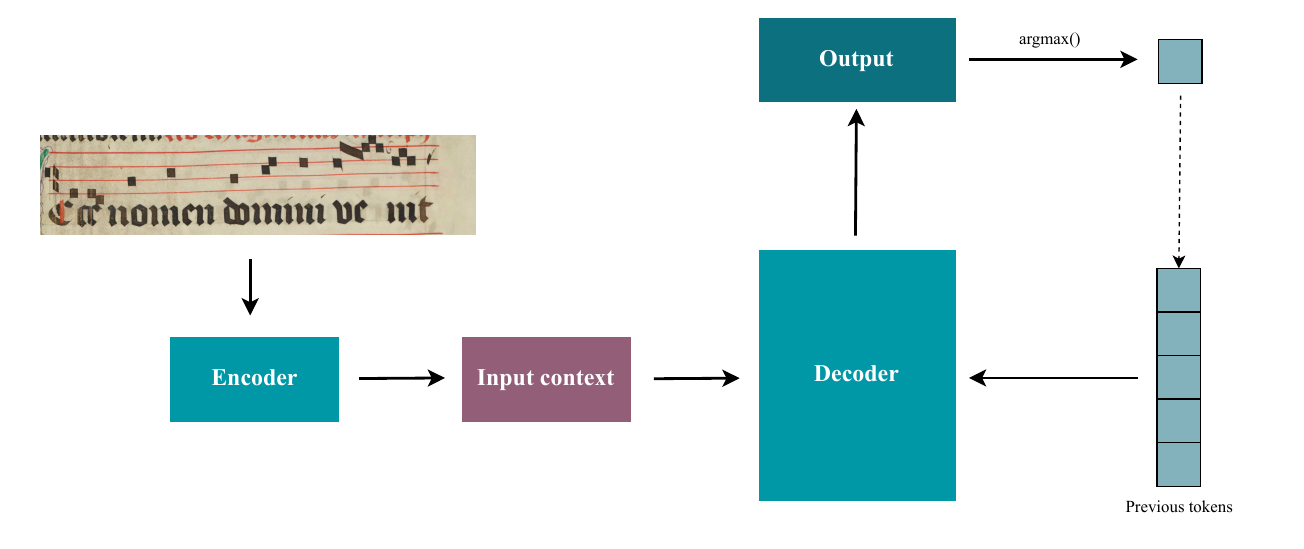}
    \caption{Example of an autoregressive language model for transcription, based on the Seq2Seq model~\cite{NIPS2014_a14ac55a}.}
    \label{fig:lm}
\end{figure}

Language modeling-based solutions have emerged as an alternative to \ac{CTC}-trained models, overcoming these limitations in both the \ac{HTR} and \ac{OMR} fields~\cite{Kang:PR:2022,Coquenet:TPAMI:2023,Singh:ICDAR:2021,Dhiaf2023,smt_toto}. These models are autoregressive end-to-end neural networks, primarily based on the Transformer architecture~\cite{Vaswani:NIPS:2017}, which are able to generate the transcription of an input image token by token. Specifically, they consist of an encoder that extracts relevant features from the input image and a decoder that generates the transcription conditionally to a given prefix. A graphical example of these systems is shown in Fig.~\ref{fig:lm}.

These models excel at learning complex reading orders in documents, thanks to their independence from sequential image features and their ability to exploit contextual relationships between sequences. We hypothesize that these architectures can learn the specific reading order of \ac{AMNLT} scores without requiring pre-processing steps. Additionally, they can leverage the contextual relationships defined by the alignment vocabulary to produce both accurate and syntactically correct outputs.

\section{Case study: Early music notation}
\label{sec:case_study}
Our case study focuses on early music notation, chosen for its significant musicological and historical interest as well as its relationship with \ac{AMNLT}. Early music predominantly features vocal compositions, making it a representative domain for addressing the challenges of transcription and alignment in vocal music scores.

In this section, we present the publicly available corpora for \ac{AMNLT}, as well as the encoding formats used to meet the challenge's requirements. Specifically, this paper features four distinct datasets: one hybrid dataset with transcriptions sourced from a well-known database, and three datasets composed of scanned books.\footnote{Here ``hybrid'' means a dataset which contains real music ground-truth, but music score images are synthetically rendered.}

\subsection{Corpora}
\begin{figure}[h]
    \centering
    \includegraphics[width=\linewidth]{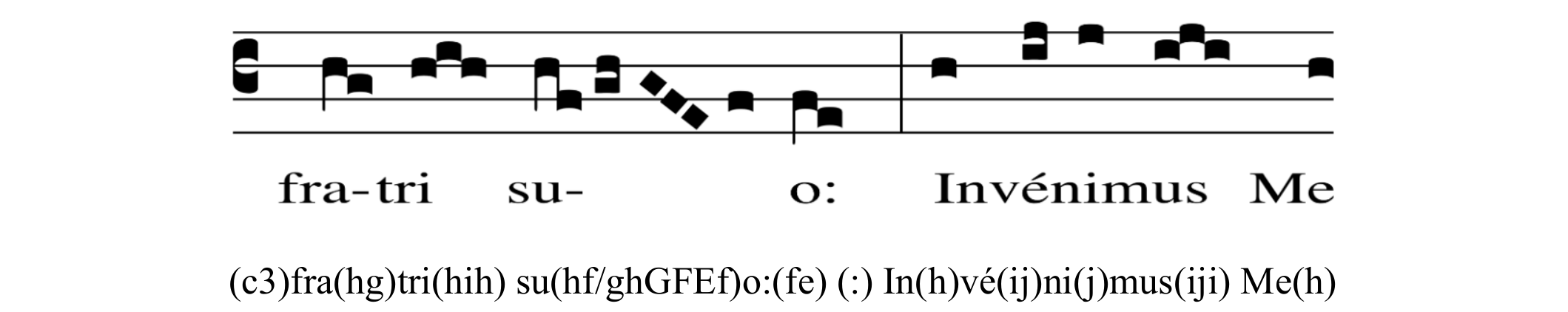}
    \caption{Sample of a system from the \GregoSynth{} dataset, with its corresponding \gabc{} transcription below.}
    \label{fig:gregosynth}
\end{figure}

The first corpus is the \GregoSynth{} dataset, a hybrid dataset generated by processing the Gregorian Chant Database.\footnote{\url{https://gregobase.selapa.net/}} This database contains nearly $20,000$ music score pages annotated in the \gabc{} encoding format, a character-based annotation standard for Gregorian chant scores. The production of this corpus followed a two-step workflow: (i) splitting the full-page \gabc{} encoding into single vocal systems and (ii) rendering the extracted samples using the GregorioTex online tool.\footnote{\url{https://gregorio-project.github.io/gregoriotex/}} Figure~\ref{fig:gregosynth} illustrates an example produced by this pipeline.

\begin{figure}[h]
    \centering
    \includegraphics[width=\linewidth]{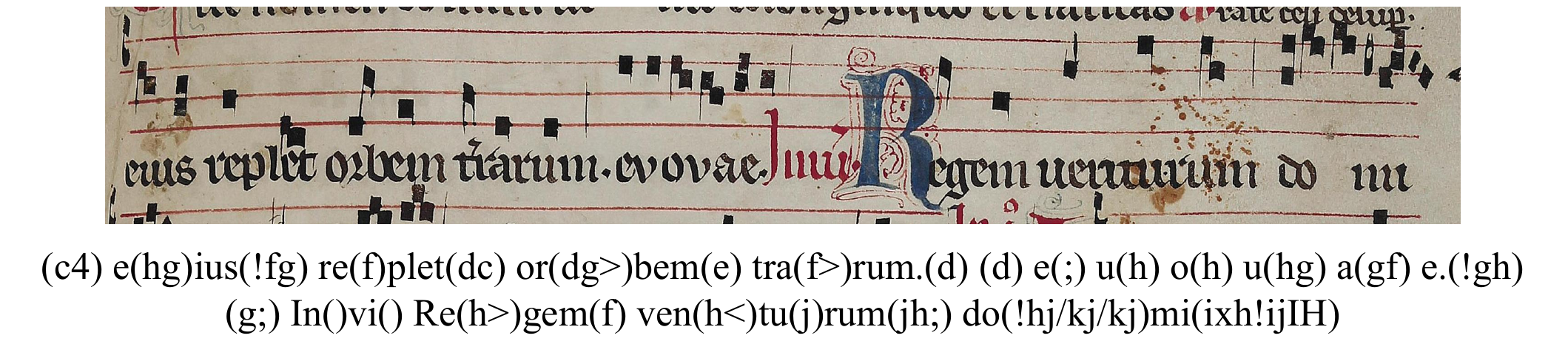}
    \caption{Sample of a system from the \Solesmes{} dataset, with its corresponding \sgabc{} transcription below.}
    \label{fig:solesmes}
\end{figure}

The second corpus is the \Solesmes{} dataset, developed within the Repertorium project, a European initiative focused on annotating unpublished Gregorian chants from the Solesmes abbey.\footnote{\url{https://repertorium.eu/}} It comprises $854$ music systems annotated in the \sgabc{} encoding language~\cite{sgabc}, which provides a more systematic and less verbose structure for encoding \gabc{} documents. Figure~\ref{fig:solesmes} presents an example from this dataset.

\begin{figure}[h]
    \centering
    \includegraphics[width=\linewidth]{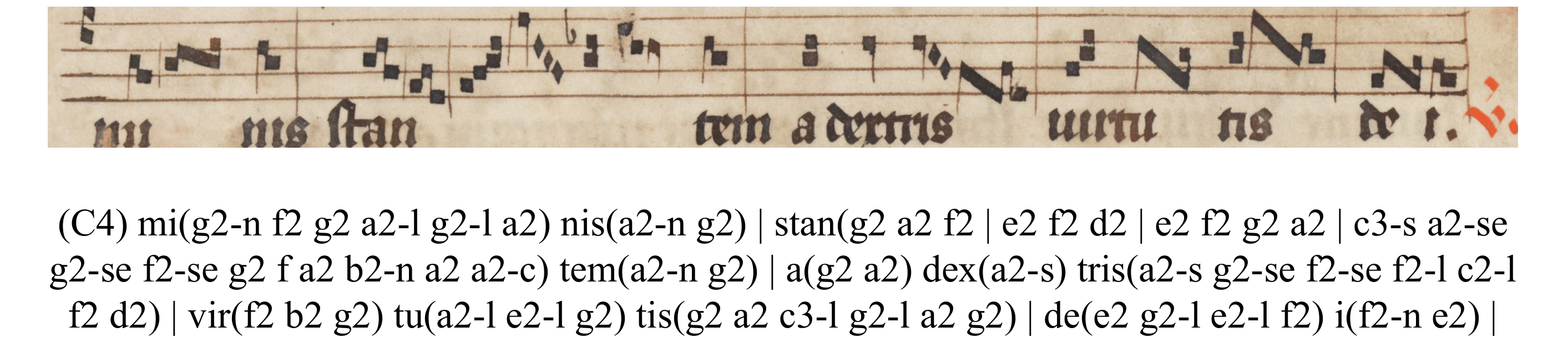}
    \caption{Sample of a system from the \Einsiedeln{} dataset, with its corresponding \pgabc{} transcription below.}
    \label{fig:einsiedeln}
\end{figure}

\begin{figure}[h]
    \centering
    \includegraphics[width=\linewidth]{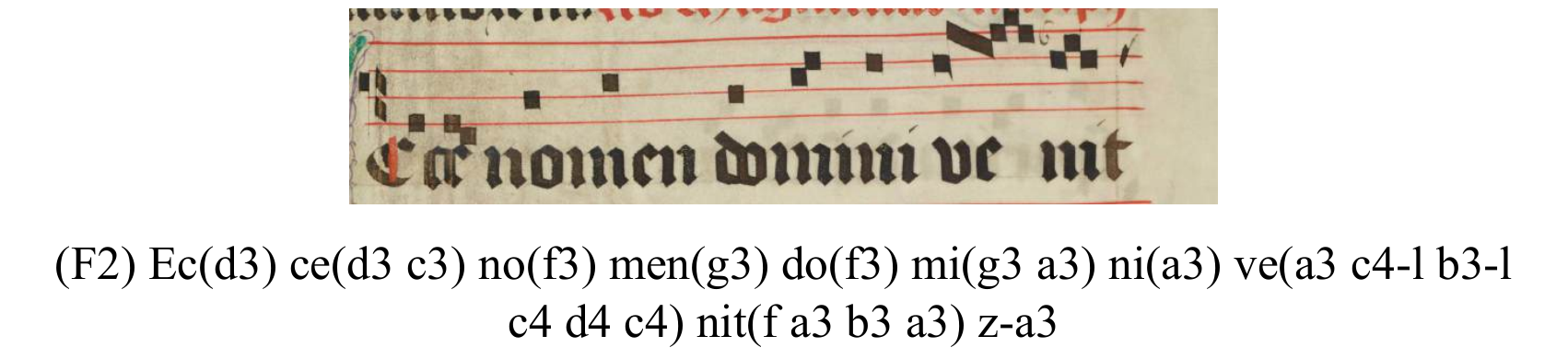}
    \caption{Sample of a system from the \Salzinnes{} dataset, with its corresponding \pgabc{} transcription below.}
    \label{fig:salzinnes}
\end{figure}

Finally, the \Einsiedeln{} and \Salzinnes{} corpora were derived from the CantusDB project.\footnote{\url{https://cantusdatabase.org/}} The \Einsiedeln{} corpus originates from a 14th-century antiphonary from the monastery of Einsiedeln, Switzerland, while the \Salzinnes{} corpus comes from a 16th-century Cistercian antiphoner from the Abbey of Salzinnes, Namur, in the Diocese of Liège. These corpora contain $1816$ and $2965$ annotated vocal excerpts, respectively, encoded in the \ac{MEI} format. Examples from the \Einsiedeln{} dataset and \Salzinnes{} dataset are shown Fig.~\ref{fig:einsiedeln} and Fig.~\ref{fig:salzinnes}, respectively.

A summary of the features of the datasets presented in this work is provided in Table~\ref{tab:corpora}.

\subsection{Output encoding adaptation}
The datasets described above, although labeled in well-known annotation formats, require preprocessing to align with the \ac{AMNLT} formulation. 

For the \GregoSynth{} and \Solesmes{} datasets, the \gabc{} standard format uses the same character set for both music notes and lyrics. While this compact vocabulary is efficient, it introduces noise during training, as the network must predict the same character for two distinct graphic symbols. To address this issue, we propose a \textit{music-aware} \gabc{} encoding. In this encoding, all characters enclosed within alignment symbols---represented by parentheses---are assigned a \textless\textit{m}\textgreater prefix. Further discussion and experimental evaluation of this encoding approach can be found in~\ref{app1}.

For the \Einsiedeln{} and \Salzinnes{} datasets, although the \ac{MEI} standard satisfies the \ac{AMNLT} requirements, it is known for being verbose. To simplify this, we adapted it into a \emph{pseudo} \gabc{} notation. Using the tree structure of \ac{MEI}, we reduced each musical note to its fundamental elements while maintaining a clear separation between lyrics and music, as required by the \ac{AMNLT} specifications. This simplified representation, referred to as \pgabc{}, is reversible back to standard \ac{MEI} through a straightforward rule-based conversion system.

\begin{figure}[h]
    \centering
    \includegraphics[width=0.8\linewidth]{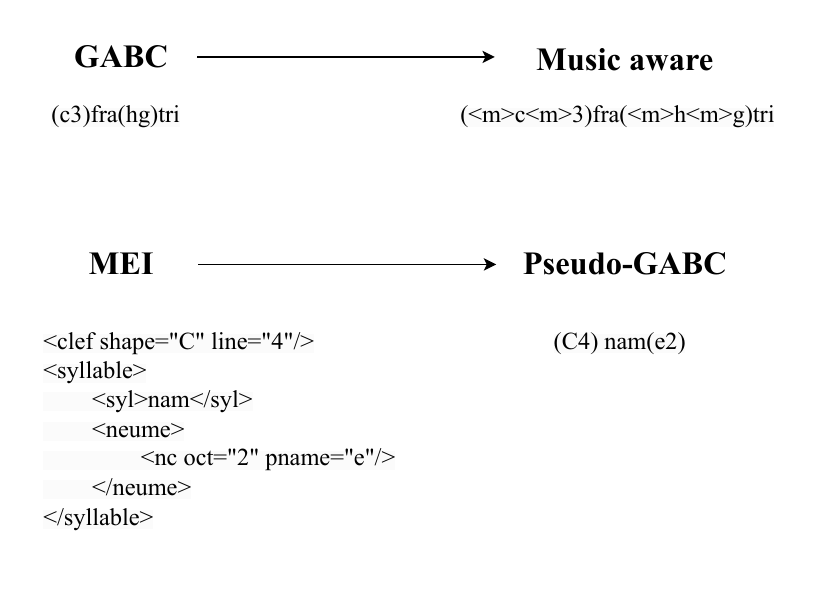}
    \caption{Example of the adaptation of the output encodings.}
    \label{fig:output_encodings}
\end{figure}

\begin{table}[!ht]
\caption{Overview of the datasets used in this work, including the number of annotated systems, unique tokens in the vocabulary, data type (real or hybrid), and the original encoding format.}
\label{tab:corpora}
\centering
\renewcommand{\arraystretch}{1.5}
\begin{adjustbox}{max width=\textwidth}
\begin{tabular}{lcccc}
\toprule[1pt]
                                & \textbf{Systems}   & \textbf{Unique tokens} & \textbf{Data type}      & \textbf{Original notation} \\
\cmidrule(lr){1-5}
\GregoSynth     & 126\,579  &  399       & Hybrid    & \gabc     \\
\Solesmes       & 854       &  137       & Real      & \sgabc    \\
\Einsiedeln     & 1\,816    &  177       & Real      & MEI       \\
\Salzinnes      & 2\,965    &  183       & Real      & MEI       \\
\bottomrule[1pt]
\end{tabular}
\end{adjustbox}
\end{table}

\section{Metrics for AMNLT}
\label{sec:metrics}
In this section, we formally define the metrics used to measure the performance of \ac{AMNLT} systems. Since we are dealing with transcription systems, we base our evaluation on a standard measurement in the \ac{OMR} and \ac{OCR} fields: the edit distance. This metric calculates the total number of editing operations required to transform a hypothesis sequence into a reference sequence. Adapting this process to the \ac{AMNLT} challenge results in three distinct metrics, described below: Music Error Rate, Character Error Rate, and Syllable Error Rate. However, it has been shown that edit distance-based metrics fail to differentiate between content errors and alignment errors, as they merge all information into a single computation~\cite{VIDAL2023109695}. To address this limitation, we introduce a new measure specific to \ac{AMNLT} that focuses on alignment accuracy: the Alignment Error Rate.

\paragraph{Music Error Rate (MER)}
\label{para:mer}
This metric evaluates the musical aspect of \ac{AMNLT}, focusing solely on music transcriptions. It calculates the edit distance at the music token level. In the \emph{music-aware} \gabc{} encoding (Fig.~\ref{fig:output_encodings}), a music token is defined as any character enclosed in parentheses and prefixed by the \( \mathcal{<} \)m\( \mathcal{>} \) tag. For evaluation purposes, the tag is ignored to avoid skewing results. In the \pgabc{} notation (see Fig.~\ref{fig:output_encodings}), a music token is any group of characters enclosed in parentheses and separated by spaces. This ensures even minor differences in musical notation are captured, providing a detailed analysis of transcription accuracy. A lower \MER{} indicates higher transcription quality for the music content.

\paragraph{Character Error Rate (CER)}
\label{para:cer}
This metric evaluates the transcription quality of the lyrics at the character level. By computing the edit distance for each individual character in the lyrics, \CER{} captures discrepancies such as spelling errors or missing characters. This fine-grained metric provides a detailed evaluation of the lyrical transcription while avoiding excessive penalization for minor deviations. A lower \CER{} signifies better transcription accuracy for the lyrics.

\paragraph{Syllable Error Rate (SylER)}
\label{para:sylER}
This metric evaluates the lyrics at the syllable level, rather than at the character level. Since lyrics in music scores are typically read syllable by syllable, \SylER{} provides a more natural and accurate assessment of lyrical transcription quality. By breaking the text into syllables, it aligns more closely with how lyrics are perceived and sung, capturing errors such as syllable merging or splitting that might be overlooked by \CER{}. A lower \SylER{} indicates better performance in transcribing lyrics at a granular, syllabic level.

\paragraph{Alignment Error Rate (AlER)}
\label{para:aler}
The \AlER{} metric is designed to evaluate alignment accuracy in \ac{AMNLT} systems. It consists of two components: 
\begin{itemize}
\item The Aligned Music \& Lyrics Error Rate (\AMLER{}), which evaluates the entire transcription, including both musical elements and lyrics, as well as their synchronization. Unlike \MER{} and \SylER{}, \AMLER{} measures how well the predicted transcription matches the ground truth in terms of both accuracy and alignment. 
\item The Baseline Word Error Rate (\bWER{}), inspired from the work of Vidal et al.~\cite{VIDAL2023109695}, which computes the content-only error while ignoring alignment.
\end{itemize}

To calculate the \AlER{}, we subtract the \bWER{} from the \AMLER{} and normalize it by the \AMLER{}:

\begin{equation}
\label{eq:aler}
\mathbf{AlER} = \frac{\mathbf{AMLER} - \mathbf{bWER}}{\mathbf{AMLER}}
\end{equation}

This provides an explicit assessment of alignment errors, indicating the proportion of the total error that can be attributed to misalignments. A lower \AlER{} reflects better alignment accuracy. Further discussion of the \AlER{} metric is provided in~\ref{app2}.

\section{Implementation details}
\label{sec:amnlt_approaches_implementation}
This section describes the implementation details of the different approaches presented in Section~\ref{sec:amnlt_approaches}.\footnote{The source code of the experiments described in this section are available at \url{https://github.com/efm18/AMNLT.git}}

\subsection{Divide \& conquer}
\label{sec:dc_implementation}
The architecture implemented for each of the two models involved in the \emph{divide \& conquer} approach is the \ac{CRNN}. This architecture implements a \ac{CNN} for feature extraction and the \ac{RNN} as a sequence processor to exploit the temporal dependencies of the previous step. The model outputs a probability posteriorgram that is converted into the output text sequence through a greedy decoding strategy. Once each separate prediction is obtained, they are combined with any of the post-aligning methods, which are objects to study in this work.

The architecture of this model is composed of four convolutional layers with 64 kernels of size \({5 \times 5}\), 64 kernels of size \( 5 \times 5\), 128 kernels of size \(3 \times 3\) and 128 kernels of size \(3 \times 3\), respectively. We consider a Leaky ReLU activation with a negative slope of \( \alpha = 0.2 \) and max-pooling stages with a size and stride factors of \( 2 \times 1\) (except for the first convolutional layer, which is \( 2 \times 2\)). The produced feature maps were introduced into the first two \ac{BLSTM} layers with 256 hidden units each and a dropout of 0.5, followed by a fully connected network with $|\Sigma^{\prime}_S|$ units, where $|\Sigma^{\prime}_S|$ is the vocabulary size of the model.


\subsection{Base holistic approach}
\label{sec:base_holistc_implementation}
The base holistic approach follows the same architecture as described in Section~\ref{sec:dc_implementation}, built on the state-of-the-art for \ac{OMR}. However, in this case, the model receives the entire music excerpt image, unlike the \emph{divide \& conquer}, which processes music and lyrics separately.

\subsection{Unfolding}
\label{sec:unfolding_implementation}
The \emph{unfolding} method implementation is based in the work of Coquenet et al.~\cite{Coquenet2021}. As described in Sect.~\ref{sec:unfolding}, the model rearranges the output feature map from the encoder---with dimensions \( c, h, w \)\footnote{The variable $c$ stands for the number of features from the last layer of the encoder, $h$ the height, rows, and $w$ the width, columns of the feature map}---by concatenating each of its rows in the form of \( c, h \times w \). Figure~\ref{fig:unfolding_stave} illustrates this process. We implement three variants of this network: (i) a \ac{FCN}, (ii) a \ac{CRNN}, and (iii) a \ac{CNNT2D}, whose architectures are explained below.

\paragraph{FCN}
\label{para:fcn}
This architecture combines convolutional and \ac{DSC}~\cite{liu2022convnet} blocks to process as input the rotated systems images and produce a probability map of output categories, which is refined by a \ac{CTC} loss function. It consists of six convolutional blocks with increasing filters, from 32 to 512, ReLU activation, batch normalization, and mixed dropout~\cite{Coquenet:ICFHR:2020}. These are followed by four \ac{DSC} blocks with 512 filters and residual connections. The decoder includes a final convolutional layer to map features to the output vocabulary.

\paragraph{CRNN}
\label{para:crnn}
This architecture implements the same convolutional encoder as in the \ac{FCN} approach, but adds a recurrent decoder to process the temporal dependencies from the extracted features. Specifically, we incorporate a \ac{BLSTM} with two layers, which outputs feature sequences, followed by a linear layer to map the features to output categories.

\paragraph{CNNT2D}
\label{para:cnnt2d}
This last implementation of the Unfolding mechanism leverages the Transformer layers for temporal dependency processing. Specifically, we replace the BLST from the \ac{CRNN} approach with a Transformer encoder. This network particularly implements a 2D Positional Encoding, which allows to grasp better spatial relationships in the feature map, which is an image by nature where elements are placed on top of each other. This has proven to give better performance in the case of \ac{AMNLT}, at least in synthetic samples~\cite{MartinezSevilla:ICDAR:2023}.

\subsection{Language modeling}
\label{sec:language_modeling}
We leverage the \ac{SMT} as the language modeling approach for \ac{AMNLT}~\cite{smt_toto}. This model is a Transformer-based end-to-end method that was implemented for transcribing complex music scores, such as those for piano. Since vocal scores can be seen graphically analogous, as mentioned in Section~\ref{sec:amnlt}, we consider this model suitable for the task of \ac{AMNLT}.

The \ac{SMT} model is built on a transformer-based image-to-sequence framework, featuring two primary components: an encoder and an autoregressive decoder. The encoder is based on a ConvNexT network~\cite{liu2022convnet}, which has shown outstanding performance in the~\ac{SMT}. The ConvNexT consists of hierarchical convolutional layers that downscale the input image, producing a feature map that captures both low-level (e.g., note shapes, staff lines) and high-level (e.g., musical symbols, textual elements) patterns. Specifically, it maintains the first three stages, reducing the input image by a factor of $16$. The output of the encoder is flattened to form a sequence that serves as input to the decoder. 2D Positional Encoding is applied to preserve spatial relationships within the score.

The decoder is a transformer-based sequence generator that uses multi-head self-attention to capture the temporal and contextual dependencies between different tokens. At each time step, the decoder predicts the next token, whether it be a note, rest, or lyric, based on the features extracted by the encoder and the sequence of previously predicted tokens.

\section{Results}
\label{sec:results}
Table~\ref{tab:results_gregosynth_solesmes} and Table~\ref{tab:results_einsiedeln_salzinnes} summarize the performance of the proposed models for \ac{AMNLT} across the four corpora presented in this work. The tables are organized by encoding type: Table~\ref{tab:results_gregosynth_solesmes} presents results for the \emph{music-aware} \gabc{} datasets---\GregoSynth{} and \Solesmes{}---while Table~\ref{tab:results_einsiedeln_salzinnes} displays results for the \pgabc{} datasets, \Einsiedeln{} and \Salzinnes{}.

The results indicate that end-to-end approaches generally outperform the \emph{divide \& conquer} baseline in alignment accuracy, as reflected by lower \AlER{} scores. However, it is essential to interpret the \AlER{} metric alongside the overall transcription quality metrics (\MER{}, \CER{}, and \SylER{}). When a model exhibits poor transcription quality (i.e., high \MER{}, \CER{}, and \SylER{}), most errors are likely content-related rather than alignment-related. 

An example of this behavior is observed in the frame-level post-alignment method within the \emph{divide \& conquer} approach. This method employs a greedy pairing strategy that discards some content when selecting the first eligible music-lyrics pair. Consequently, certain lyrics are duplicated or omitted, resulting in content issues being the primary contributor to the overall error. This phenomenon is evident in the \Solesmes{} dataset, where the \emph{divide \& conquer} approach with frame-level post-alignment achieves the best \AlER{} score but ranks as the second-worst method in terms of \AMLER{}. This example highlights the importance of jointly analyzing \AlER{} and \AMLER{} to gain a comprehensive understanding of model performance. Considering only alignment accuracy without evaluating transcription quality may lead to misleading conclusions about the effectiveness of a given method.


\begin{table*}[!h]
\caption{Performance results for the four approaches applied to the \GregoSynth{} and \Solesmes{} datasets. The table reports metrics for music transcription (\MER{}), lyrics transcription (\CER{} and \SylER{}), and alignment accuracy (\AMLER{} and \AlER{}). Results are organized by approach (\textit{Divide \& Conquer}, \textit{Base Holistic}, \textit{Unfolding}, and \textit{Language Modeling}) and implementation strategy for each method. The best-performing values for \MER{}, \CER{}, \SylER{}, and \AMLER{} are highlighted in bold.}
    \label{tab:results_gregosynth_solesmes}
    \centering
    \resizebox{\textwidth}{!}{
    \begin{tabular}{lllcccccccccccccc}
        \toprule[1.5pt]
        \multicolumn{3}{l}{\textbf{\textit{Approach}}} && \multicolumn{5}{c}{\GregoSynth} && \multicolumn{5}{c}{\Solesmes} \\
        \cmidrule(lr){5-9} \cmidrule(lr){11-15}
        & \multicolumn{2}{l}{\textbf{\textit{Implementation}}} && \MER & \CER & \SylER & \AMLER & \AlER && \MER & \CER & \SylER & \AMLER & \AlER \\
        
        \cmidrule(lr){1-15}
        \multicolumn{3}{l}{\textbf{\textit{Divide \& Conquer}}} &&&&&& \\

        \phantom{....} & \multicolumn{2}{l}{\textbf{\textit{CRNN-CTC}}} &&&&&& \\
            &\phantom{....}& Syllable && \multirow{2}{*}{2.79} & \multirow{2}{*}{\textbf{4.54}} & \multirow{2}{*}{\textbf{8.66}} & 13.40 & 0.59 && \multirow{2}{*}{\textbf{17.72}} & \multirow{2}{*}{\textbf{8.86}} & \multirow{2}{*}{\textbf{20.84}} & 23.08 & 0.53 \\
            && \ac{CTC} frames && \phantom{0} & \phantom{0} & \phantom{0} & 22.29 & 0.30 && \phantom{0} & \phantom{0} & \phantom{0} & 57.48 & 0.19 \\

        \cmidrule(lr){1-15}
        \multicolumn{3}{l}{\textbf{\textit{Base Holistic}}} &&&&&& \\

        & \multicolumn{2}{l}{\textbf{\textit{CRNN-CTC}}} && 59.99 & 89.29 & 98.91 & 62.47 & 0.10 && 24.91 & 38.36 & 73.59 & 29.71 & 0.28 \\
        
        \cmidrule(lr){1-15}
        \multicolumn{3}{l}{\textbf{\textit{Unfolding}}} &&&&&& \\

        & \multicolumn{2}{l}{\textbf{\textit{FCN}}} && \phantom{0} 3.82 & 16.05 & 37.44 & 8.64 & 0.09 &&  25.14 & 40.51 & 88.62 & 21.08 & 0.38 \\
        & \multicolumn{2}{l}{\textbf{\textit{CRNN}}} && \phantom{0} 4.00 & 10.77 & 25.11 & 5.85 & 0.04 &&  18.20 & 19.11 & 38.88 & \textbf{20.98} & 0.38 \\
        & \multicolumn{2}{l}{\textbf{\textit{CNNT2D}}} && \phantom{0} 22.70 & 50.03 & 89.93 & 31.28 & 0.13 &&  64.42 & 88.97 & 97.32 & 71.82 & 0.08 \\

        \cmidrule(lr){1-17}
        \multicolumn{3}{l}{\textbf{\textit{Language Modeling}}} &&&&&& \\

        & \multicolumn{2}{l}{\textbf{\textit{SMT}}} && \phantom{0} \textbf{2.26} & 6.39 & 15.82 & \textbf{2.93} & 0.09 &&  35.37 & 52.87 & 91.00 & 41.19 & 0.39 \\
        \bottomrule[1.5pt]
\end{tabular}}
\end{table*}

The results indicate that the \emph{divide \& conquer} baseline approach excels in transcription tasks, achieving the best performance in almost all datasets. Notable exceptions include the language modeling approach, which achieves the best \MER{} in the \GregoSynth{} dataset, and the \SylER{} in the \Salzinnes{} corpus. However, the \emph{divide \& conquer} method performs poorly in alignment tasks, making it the worst approach for this critical aspect of \ac{AMNLT}. 

Among the proposed end-to-end approaches, the language modeling method emerges as the best-performing strategy, particularly when sufficient training data is available for convergence. On average, this method achieves a 42.3\% improvement in \AMLER{} and a 57.46\% improvement in \AlER{} compared to the best \emph{divide \& conquer} result, which is the syllable post-alignment method. The second-best approach is the unfolding method with a \ac{CRNN} architecture, achieving an average improvement of 40.02\% in \AMLER{} and 61.90\% in \AlER{}.

The base holistic approach demonstrates that even with the same architecture used in the \emph{divide \& conquer} baseline, alignment performance improves. This foundational improvement indicates the potential of end-to-end strategies to address the \ac{AMNLT} challenge. However, this approach struggles to correlate music and lyrics effectively under standard CTC training, leading to poor performance in datasets such as \GregoSynth{} and \Solesmes{}, as shown in Table~\ref{tab:results_gregosynth_solesmes}. These limitations are addressed by the more advanced end-to-end proposals.

The unfolding approach offers notable improvements over the base holistic strategy. Specifically, the \ac{CRNN} implementation consistently outperforms the \ac{FCN} and \ac{CNNT2D} variants on average. Among the individual results, the most significant improvements are observed in the \Einsiedeln{} corpus, where the unfolding \ac{CRNN} achieves a 77.30\% improvement in \AMLER{} and a 22.22\% improvement in \AlER{} compared to the baseline approach.

\begin{table*}[!h]
    \caption{Performance results for the four approaches applied to the \Einsiedeln{} and \Salzinnes{} datasets. The table reports metrics for music transcription (\MER{}), lyrics transcription (\CER{} and \SylER{}), and alignment accuracy (\AMLER{} and \AlER{}). Results are organized by approach (\textit{Divide \& Conquer}, \textit{Base Holistic}, \textit{Unfolding}, and \textit{Language Modeling}) and implementation strategy for each method. The best-performing values for \MER{}, \CER{}, \SylER{}, and \AMLER{} are highlighted in bold.}
    \label{tab:results_einsiedeln_salzinnes}
    \centering
    \resizebox{\textwidth}{!}{
    \begin{tabular}{lllcccccccccccccc}
        \toprule[1.5pt]
        \multicolumn{3}{l}{\textbf{\textit{Approach}}} && \multicolumn{5}{c}{\Einsiedeln} && \multicolumn{5}{c}{\Salzinnes} \\
        \cmidrule(lr){5-9} \cmidrule(lr){11-15}
        & \multicolumn{2}{l}{\textbf{\textit{Implementation}}} && \MER & \CER & \SylER & \AMLER & \AlER && \MER & \CER & \SylER & \AMLER & \AlER \\
        
        \cmidrule(lr){1-15}
        \multicolumn{3}{l}{\textbf{\textit{Divide \& Conquer}}} &&&&&& \\

        \phantom{....} & \multicolumn{2}{l}{\textbf{\textit{CRNN-CTC}}} &&&&&&\\
            &\phantom{....}& Syllable && \multirow{2}{*}{\textbf{11.36}} & \multirow{2}{*}{\textbf{5.47}} & \multirow{2}{*}{\textbf{12.68}} & 35.08 & 0.54 && \multirow{2}{*}{22.44} & \multirow{2}{*}{\textbf{2.84}} & \multirow{2}{*}{7.98} & 14.70 & 0.33 \\
            && \ac{CTC} frames && \phantom{0} & \phantom{0} & \phantom{0} & 90.94 & 0.11 && \phantom{0} & \phantom{0} & \phantom{0} & 87.76 & 0.17 \\

        \cmidrule(lr){1-15}
        \multicolumn{3}{l}{\textbf{\textit{Base Holistic}}} &&&&&& \\

        & \multicolumn{2}{l}{\textbf{\textit{CRNN-CTC}}} && 11.82 & 9.95 & 23.06 & 10.21 & 0.15 &&  19.9 & 5.88 & 13.93 & 10.89 & 0.13 \\
        
        \cmidrule(lr){1-15}
        \multicolumn{3}{l}{\textbf{\textit{Unfolding}}} &&&&&& \\

        & \multicolumn{2}{l}{\textbf{\textit{FCN}}} && 30.41 & 31.96 & 65.28 & 26.74 & 0.20 && 67.25 & 62.83 & 98.35 & 54.49 & 0.24 \\
        & \multicolumn{2}{l}{\textbf{\textit{CRNN}}} && 14.26 & 7.96 & 18.6 & 10.43 & 0.17 && 20.37 & 6.23 & 14.59 & 11.12 & 0.14 \\
        & \multicolumn{2}{l}{\textbf{\textit{CNNT2D}}} && 79.60 & 61.54 & 98.90 & 56.18 & 0.10 && 80.24 & 16.33 & 39.65 & 43.49 & 0.12 \\

        \cmidrule(lr){1-17}
        \multicolumn{3}{l}{\textbf{\textit{Language Modeling}}} &&&&&& \\

        & \multicolumn{2}{l}{\textbf{\textit{SMT}}} && 14.05 & 8.45 & 15.84 & \textbf{10.78} & 0.21 && \textbf{13.73} & 3.69 & \textbf{7.80} & \textbf{7.32} & 0.14 \\
        \bottomrule[1.5pt]
\end{tabular}}
\end{table*}

Concerning the language modeling approach, it is important to note that its average result is negatively impacted by poor performance on the \Solesmes{} dataset. This is primarily due to the limited number of samples in \Solesmes{} (see Table~\ref{tab:corpora}), which might be insufficient for the \ac{SMT} model to converge. However, when provided with enough data, this approach achieves the best \AMLER{} and \AlER{} values. For example, in the \GregoSynth{} dataset, it achieves an \AMLER{} of 2.93\% and an \AlER{} of 0.09, and in the \Salzinnes{} dataset, it reports an \AMLER{} of 7.32\% and an \AlER{} of 0.14.

Overall, these results demonstrate that end-to-end methods are generally superior for \ac{AMNLT} compared to the baseline \emph{divide \& conquer} approaches. End-to-end models produce more meaningful results, as they integrate transcription and alignment within a single framework. Let us recall that alignment is a key concept for generating interpretable and processable results within the musical context.

Among the end-to-end approaches, the language modeling method delivers the best overall performance, although it requires a sufficient quantity of training data to achieve this. When this condition is not met, the unfolding approach with recurrent sequence processing (\ac{CRNN}) provides the best alternative, particularly for datasets like \Solesmes{}. 

Our findings highlight a trade-off between transcription precision and alignment accuracy. In this comparison, the language modeling approach achieves the best balance, producing fully aligned results with only a slight performance drop compared to the \emph{divide \& conquer} method.

\section{Conclusions}
\label{sec:conclusions}
In this paper, we provide a foundational framework for \ac{AMNLT} for the first time. This task integrates music and lyrics transcription while explicitly considering their synchronization during interpretation, referred here to as alignment. 

We have formally defined and formulated the challenge, analyzed existing methods, and proposed several approaches. Specifically, we categorized these methods into two families: \emph{divide \& conquer}, following traditional state-of-the-art pipelines, and end-to-end approaches, which generate the complete transcription of a score in a single step. For the end-to-end family, we proposed three specific methods: direct transcription, unfolding, and language modeling.

Our study focuses on the transcription of medieval chants, a domain of particular interest for \ac{AMNLT}. To support this research, we introduced four publicly available benchmark datasets: \GregoSynth{}, \Solesmes{}, \Einsiedeln{}, and \Salzinnes{}. Additionally, we proposed two novel metrics, \AMLER{} and \AlER{}, to assess both transcription quality and alignment precision. 

The experimental results demonstrate that end-to-end approaches are generally more effective for \ac{AMNLT}, providing strong transcription quality with meaningful alignments. Among these, language models outperform other methods, achieving comparable performance to the baseline \emph{divide \& conquer} approach. However, \emph{divide \& conquer} methods still excel in transcription quality due to their ability to focus on music and lyrics independently.

Our results establish a foundation for future research on \ac{AMNLT} and highlight a trade-off between transcription precision and alignment quality. Several directions for future research emerge from this work. One critical area is the improvement of end-to-end methods' pure transcription accuracy, which still lags behind \emph{divide \& conquer} approaches. Special attention should be given to language models, where more data-efficient strategies could not only improve performance but also enable effective transcription of smaller corpora. Another promising avenue is the development of improved post-alignment methods for \emph{divide \& conquer} approaches. Addressing information loss during training and alignment could result in hybrid methods that leverage the strengths of both approaches---combining high transcription quality with precise alignment.

\section*{Acknowledgments}
This paper is part of REPERTORIUM project, funded by the European Union’s Horizon Europe programme under grant agreement No 101095065. The second autor is supported by grant ACIF/2021/356 from the ``Programa I+D+i de la Generalitat Valenciana''.

\bibliographystyle{splncs04}
\bibliography{bibliography}

\newpage
\appendix

\section{\gabc{} encoding}
\label{app1}

In this paper, we resort to a variant of the \gabc{} music encoding format to experiment with the \GregoSynth{} and the \Solesmes{} databases. Although the performance results are very positive, this topic might need an extended analysis and discussion of the decisions taken. 

\begin{figure}[h]
    \centering
    \includegraphics[width=\linewidth]{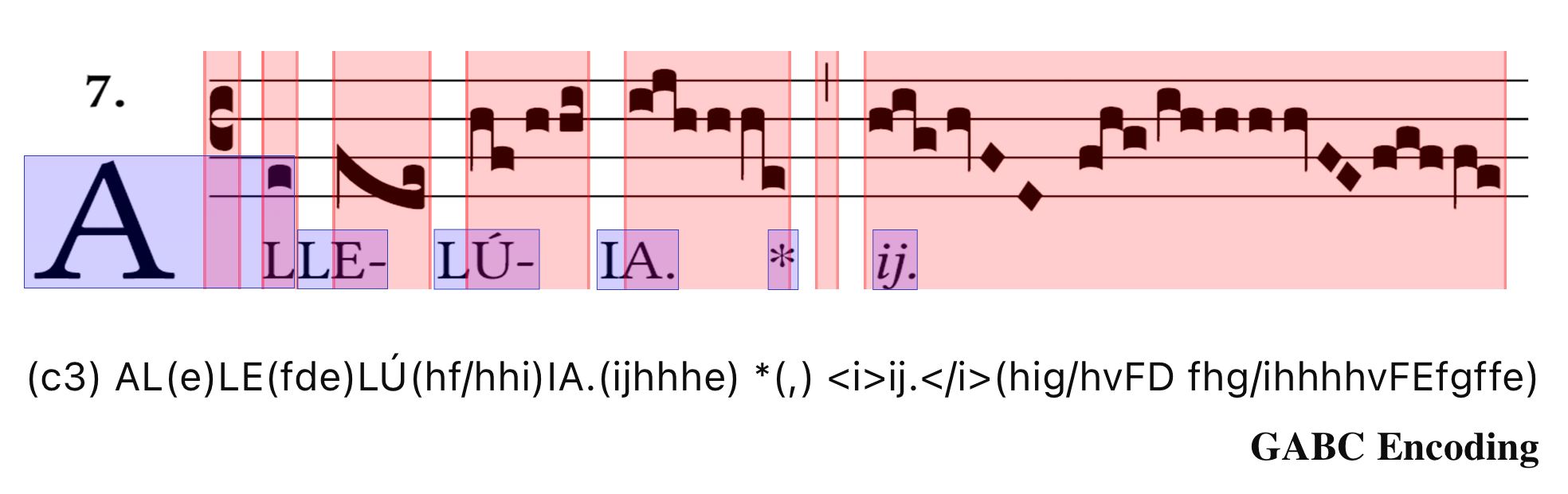}
    \caption{Example of the alignment between a music-symbol sequence and a text sequence in a Gregorian melody fragment. Red boxes refer in a pixel-wise viewpoint to the area of a music symbol inside the image, whereas blue boxes represent the same for the syllables. The \gabc{} encoding is also presented below, where music notation is encapsulated in between parentheses, and lyrics are left outside them.}
    \label{fig:gabc_simple_example}
\end{figure}

\gabc{} is an ASCII-based music notation language, which effectively annotates Gregorian chants by representing the music, lyrics, and their alignment through a single char set. \gabc{} encapsulates all the music elements between parentheses after each syllable. An example is shown in Fig.\ref{fig:gabc_simple_example}.

This format was selected over others, such as the Volpiano encoding \cite{Helsen_Lacoste_2011}, due to its comprehensive representation of both the melody and the text, including their alignment. Although Volpiano is a widely recognized standard in the Cantus database for encoding melodies \cite{lacoste2012cantus}, it is not able to represent the text in syllables, which are the main unit used for aligning music notation and lyrics. This disadvantage makes \gabc{} more suitable for the needs of \ac{AMNLT}, as it provides an integrated approach to handling both musical notes and lyrics.

Despite its goodness in comparison with other encodings, it is still not directly suitable for \ac{AMNLT}. Specifically, \gabc{} uses the same charset to represent music and lyrics. That is, a single character and a music note are annotated with the same token. This could potentially lead to training inconsistencies because of the noisy data. This issue is solved by creating the \emph{music-aware} encoding, which is formally presented in this paper. In this encoding, each music token is preceded by a \( \mathcal{<} \)m\( \mathcal{>} \) tag, thus differentiating these characters from the lyrics ones.

Our decision is assessed empirically. Specifically, we conducted an experiment with the \Solesmes{} and the \GregoSynth{} datasets, where we compare the performance of the proposed approaches in plain \gabc{} and then with the \emph{music-aware} \gabc{} format. The results, reported in Table~\ref{tab:muaw_vs_plain}, show that \emph{music-aware} \gabc{} generally outperforms raw \gabc{}. There are only two cases where this tendency is not followed. These are the \ac{CNNT2D} architecture of the unfolding approach and in the base holistic approach of the \Solesmes{} dataset.

\begin{table}[!ht]
\caption{Comparison between music-aware encoding and plain \gabc{} with \GregoSynth{} and \Solesmes{} dataset, in terms of \AMLER{} (\%).}
\label{tab:muaw_vs_plain}
\centering
\renewcommand{\arraystretch}{1.25}
\begin{tabular}{llccccc}
\toprule[1pt]
                            && \multicolumn{2}{c}{\GregoSynth}   && \multicolumn{2}{c}{\Solesmes} \\
\cmidrule(lr){3-4}  \cmidrule(lr){6-7}
                            && Plain & Music aware               && Plain    & Music aware   \\
\cmidrule(lr){1-7}
\multicolumn{2}{l}{\textbf{Base holistic}} &&&&& \\
\phantom{....} & \textit{CRNN-CTC} & 68.90 & 62.47 && 27.26 & 29.71 \\
\multicolumn{2}{l}{\textbf{Unfolding}} &&&&& \\
\phantom{....} & \textit{FCN} & 11.96 & 8.64 && 40.50 & 34.43 \\
\phantom{....} & \textit{CRNN} & 8.41 & 5.85 && 23.06 & 20.67 \\
\phantom{....} & \textit{CNNT2D} & 29.85 & 31.28 && 22.47 & 76.78 \\
\multicolumn{2}{l}{\textbf{Language modeling}} &&&&& \\
\phantom{....} & \textit{SMT} & 3.12 & 2.93 && 63.05 & 43.09 \\
\bottomrule[1pt]
\end{tabular}
\end{table}

\newpage
\section{In-depth \AlER{} analysis}
\label{app2}
To provide a deeper understanding of the \AlER, it is essential to break down how the various components contribute to this metric and to clarify its computation using examples.

The \AlER{} is designed to isolate alignment errors from content-related errors in music notation and lyrics transcription. This section elaborates on how this distinction is done through the use of two separate metrics: \AMLER{} and \bWER{}.

\subsection{\bWER: Focusing on Content Errors}
The \textit{Baseline Word Error Rate} (\bWER{}) is a metric proposed to measure the errors between two strings regardless of the reading order between the words~\cite{VIDAL2023109695}.

The \bWER{} essentially works as a content accuracy check, counting only the discrepancies between the actual musical notes and lyrics and the predicted output, but ignoring the order in which they appear. Therefore, \bWER{} is solely concerned with \emph{which} tokens were predicted and ignores \emph{where} they were placed in the sequence.

For example, if the ground truth contains the tokens \texttt{A B C} and the prediction contains \texttt{C A B}, the \bWER{} reports no error, as all the tokens match.

\subsection{\AMLER: Measuring Both Content and Alignment}
In contrast to \bWER{}, the \AMLER{} metric accounts for both content and alignment errors. This measure evaluates how similar the prediction and the ground truth are, comparing the sequences token by token in their given reading order. As the tokens are compared with the ones of their same position index in the string, \AMLER{} implicitly combines the token-level accuracy of music and lyrics with the possible aligning mismatches that could have been produced, in the same way as traditional Word Error Rate is computed~\cite{VIDAL2023109695}.

Continuing from the previous example, if the ground truth is \texttt{A B C} and the prediction is \texttt{C A B}, \AMLER{} would flag errors because, although the content set is correct, the order is not.

\subsection{Isolating alignment errors: the role of \AlER}
One important aspect when assessing \ac{AMNLT} performance is to determine how well the model is accurate at aligning music and lyrics. We have, an all-in error rate (\AMLER{}) and a content-only error rate (\bWER{}). Therefore, we propose the \AlER{} metric as the subtraction between \AMLER{} and the \bWER{}, the same way as it is proposed in the $\Delta$ metric in the work of Vidal et al.~\cite{VIDAL2023109695} for evaluation of text recognition at page level.

Mathematically, the \AlER{} is a calculated as:
\begin{equation}
\mathbf{AlER} = \frac{\mathbf{AMLER} - \mathbf{bWER}}{\mathbf{AMLER}}
\end{equation}

The result is the percentage of the total error that stems from alignment issues. When \AlER{} is high, it indicates that the bulk of the errors in the transcription are mainly because of improper synchronization between music and lyrics, while a low \AlER{} suggests that most errors are related to content inaccuracies. This metric, however, should be only taken into account in the cases where the model performs correctly. If the model produces a low transcription accuracy, \AlER{} is very likely to report low results, as the primary source of errors are from content. \AlER{}, therefore, must be always interpreted along with the rest of the \ac{AMNLT} metrics.

\subsection{Examples}
\begin{figure}[h]
    \centering
    \includegraphics[width=0.5\textwidth]{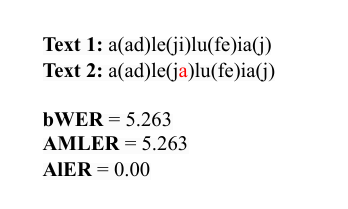}
    \caption{An example of content error. The predicted tokens contain extra or incorrect content compared to the ground truth, but the alignment is correct.}
    \label{fig:aler_0_error}
\end{figure}

\begin{figure}[h]
    \centering
    \includegraphics[width=0.5\textwidth]{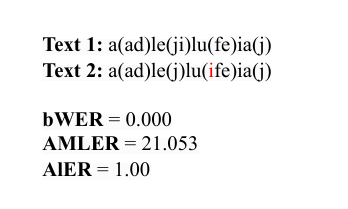}
    \caption{An example of alignment error. The predicted content matches the ground truth perfectly, but the order of tokens is incorrect.}
    \label{fig:aler_100_error}
\end{figure}

To better illustrate the elaboration above, we present two cases in Fig.~\ref{fig:aler_0_error} and~\ref{fig:aler_100_error}.

In Fig.~\ref{fig:aler_0_error}, we observe a scenario where the predicted sequence has additional tokens not present in the ground truth. This discrepancy is purely a content error, meaning the \bWER{} will be high, but the \AlER{} will be low or zero, as there is no misalignment to account for. The \AMLER{} captures the overall error, but since the misalignment is not present, the \AlER{} reflects that only content inaccuracies are affecting the transcription.

In Fig.~\ref{fig:aler_100_error}, all the content is correct and matches the ground truth. However, the predicted tokens are out of order, which represents a misalignment. Here, the \bWER{} will report a low (or zero) error since all tokens are present and correct, but the \AMLER{} will show a higher error due to the misalignment. The difference between \AMLER{} and \bWER{} will be substantial, and the \AlER{} will reflect the alignment issue as the primary source of error.

By examining these two examples, we observe how \AlER{} isolates the alignment errors, providing a clearer picture of the transcription’s quality in terms of synchronization between music and lyrics.

\end{document}